\documentclass[a4paper,11pt]{article}
\usepackage{epsfig, psfrag, amssymb, amsbsy, amsmath, amsthm}
\usepackage[table]{xcolor}
\title{Charge Scheduling of an Energy Storage System under Time-of-use Pricing and a Demand Charge}
\author{{\sc \small Yourim Yoon$^1$ and Yong-Hyuk Kim$^2$\thanks{Corresponding author}}\\
{\small $^1$ Department of Computer Engineering, College of Information Technology, Gachon University}\\
{\small 1342 Seongnamdaero, Sujeong-gu, Seongnam-si, Gyeonggi-do 461-701, Republic of Korea}\\
{\small Email: {\tt yryoon@gachon.ac.kr}}\\
{\small $^2$ Department of Computer Science \& Engineering, Kwangwoon University}\\
{\small 20 Kwangwoon-ro, Nowon-gu, Seoul 139-701, Republic of Korea}\\
{\small Email: {\tt yhdfly@kw.ac.kr}}
}

\begin{document}

\maketitle
\begin{abstract}
A real-coded genetic algorithm is used to schedule the charging of an energy storage system (ESS),
operated in tandem with renewable power by an electricity consumer who is subject to time-of-use pricing and a demand charge.
Simulations based on load and generation profiles of typical residential customers show that
an ESS scheduled by our algorithm can reduce electricity costs by approximately 17\%,
compared to a system without an ESS, and by 8\% compared to a scheduling algorithm based on net power.
\end{abstract}


\section{Introduction}

An energy storage system (ESS) is a system that is capable of absorbing energy,
storing it for a period of time, and then returning it for use.
In an electrical grid, an ESS can be used to match supply and demand.
The ESS is charged when demand is low and discharged when demand is high.
Thus, the overall energy efficiency of a system is improved, and the energy flow from the electrical grid
connected to the system is stabilized.
Reliability is a key issue in the effective use of renewable energy and in smart grids,
and thus the demand for ESSs is increasing \cite{roberts2011role}.

An ESS acts as a buffer between a generator and its load.
Renewable energy sources often generate power during off-peak periods or 
when demand for energy is low.
ESSs enable better integration of renewable energy sources into an electrical grid
by (time-shifting) the generated power and smoothing out spikes in demand.
Power producers can benefit from a more predictable generation requirement, which can improve revenue.
Installing an ESS can enable industrial, commercial, or residential end-users
to improve the quality and reliability of
their power supply and to reduce their electricity costs,
and can act as a back-up power source \cite{barton2004energy,smith2008advancement}.

Dynamic pricing of electricity is being facilitated by new technologies such as smart meters.
A form of dynamic pricing
that is being adopted in many areas is known as time-of-use (TOU)
pricing, in which electricity prices are set for a fixed period. 
Energy providers use TOU pricing to drive down demand at peak periods
by using high prices to influence customers' consumption rather than more invasive controls such as
dynamic or passive demand response mechanisms,
or even power cuts \cite{sanghvi1989flexible,sheen1994time}.
Typically TOU prices do not change more than twice a year, but
a TOU tariff is likely to have two or
three price levels (e.g., `off-peak', `mid-peak', and `on-peak')
where the price is determined by the time of day.
Customers can be expected to vary their usage in response to this price information
and manage their energy costs by shifting their usage to a lower cost period.
ESSs will play an important role in residential areas with a dynamic pricing policy.
By storing energy during low off-peak price periods and using the stored energy
when the price is high,
consumers can avoid paying high rates.

In addition to charges based on usage, an electricity bill
may include a {\em demand charge}, which is determined by
the maximum energy capacity available to a customer, whether or not it is actually used.
The demand charge is billed as a fixed rate that is calculated on a per kW basis.
This charge is based on the premise that commercial customers and other large users
should pay a share of the infrastructure
costs associated with the maintenance of capacity \cite{neufeld1987price}.
We will consider both TOU pricing and demand charges \cite{taylor1986residential}.

Many problems related to the scheduling of the charging and discharging of an ESS
have been studied recently
\cite{grillo2012optimal, hwangbo2013application, nguyen2014optimal, nottrott2013energy, squartini2013optimization}.
Various optimization techniques can be applied to the operation of ESSs.
The most frequently used method is dynamic programming, which was used by
Maly and Kwan \cite{Maly95}. They tried to
minimize electricity cost for an ESS with a given battery capacity,
without unnecessarily reducing battery life.
Ven et al.\ \cite{Ven11} aimed to minimize the capital cost of an ESS subject to 
user demand and prices, as a Markov decision process, which can be solved using dynamic programming.
Koutsopoulos et al.\ \cite{Koutsopoulos11} addressed the optimal ESS control problem from the point of view of a utility operator,
and solved the off-line problem over a finite period by dynamic programming.
Romaus et al.\ \cite{Romaus10} investigated stochastic dynamic programming for
energy management of a hybrid ESS for electric vehicles.
They aimed to control the power flow to the ESS online, while
taking into account the stochastic influences of traffic and the driver.
Huang and Liu \cite{Huang11} applied adaptive dynamic programming
to the management of a residential ESS,
with an emphasis on domestic electricity storage systems.
Their scheme was designed to learn during operation
as the environment of the ESS changes unpredictably.

There have also been a number of studies using other scheduling methods.
Youn and Cho \cite{Youn09} used linear programming to pursue
optimal operation of an energy storage unit
installed in a small power station.
Hu et al.\ \cite{Hu10} used sequential quadratic programming to operate on
ESS under real-time changes to the electricity price, so as to maximize profits.
 Non-linear programming techniques were adopted by Rupanagunta et al.\ \cite{Rupanagunta95} to design an optimal controller
for charge and discharge processes in ESSs,
with the objective of minimizing
the operating costs of the storage facility.
 Yoo et al.\ \cite{Yoo12} used a Kalman filter to increase predictability
in controlling the power flows between the components
of an energy management system for a grid-connected residential
photovoltaic (PV) system combined with an ESS under critical peak pricing.
 Lee \cite{Lee07} used multi-pass iteration particle swarm optimization
to determine the operating schedule of an ESS for an industrial TOU-rate
user who is also operating wind turbine generators.
Gallo et al.\ \cite{Gallo13} used a hybrid optimization technique
to determine values of the battery parameters required 
for an ESS operated by a smart grid management system.
Their method combines stochastic and deterministic elements within a computationally efficient algorithm. 

In this paper we describe a real-coded genetic algorithm (RCGA) for scheduling
ESS charging and discharging.
Genetic algorithms (GAs) were used by Monteiro et al.\ \cite{Monteiro13}
for short-term forecasting
 of the energy output of a PV plant.
They applied data mining techniques to historical forecasts of weather variables.
The GA was used to make spot forecasts of power output from PV plants.
We use an RCGA to schedule ESS operations under TOU pricing with a demand charge,
when a supply of renewable energy, wind or solar energy, is available.

The remainder of the paper is organized as follows:
In Section~\ref{sec:problem} we describe
the ESS scheduling problem under TOU pricing with a demand charge,
when renewable electricity is available.
In Section \ref{sec:rcga} we describe an RCGA that addresses this problem.
In Section \ref{sec:results} we present the simulation results,
and draw conclusions in Section~\ref{sec:conclusion}.

\section{ESS Scheduling Problem under TOU Pricing with a Demand Charge}
\label{sec:problem}

The formulation of our problem is similar to that of
Lee \cite{Lee07}, but we aim to optimize a daily, rather than a monthly, bill.
Other studies have dealt with optimization problems under TOU pricing.
Cao et al.\ \cite{cao2012optimized} proposed an intelligent method to control EV charging loads in response to TOU
price in a regulated market.
Lee and Chen \cite{lee1995determination} formulated the problem of determining the optimal contract
capacities and optimal sizes of ESSs for customers using a TOU rate.

\begin{table}[tb!]
\caption{Notation and expressions used in this study.\label{tab:notation}}
{\small
\begin{center}
\begin{tabular}{|l|l|} \hline
 Notation / Expression            &  Meaning \\ \hline \hline
 $l_i$                            &  load during time interval $i$ \\ 
 $g_i$                            &  energy generated during time interval $i$ \\ 
 $x_i$                            &  residual energy in the battery at the end of time interval $i$ \\ 
 $x_i - x_{i-1}$                  &  energy supplied to the battery during time interval $i$ \\ 
 $x_i - x_{i-1} +l_i - g_i$       &  net energy drawn from the grid during time interval $i$ \\ 
 $p_i$                            &  energy price set for time interval $i$ \\ 
 $p^*$                            &  fixed price \\ 
 $(x_i - x_{i-1} +l_i - g_i)p_i$  &  cost of energy over time interval $i$ \\ 
 $T$                              &  number of time intervals $= 24$ \\ 
 $C$                              &  battery capacity \\ 
 $P_c$                            &  battery charge power \\ 
 $P_d$                            &  battery discharge power \\ 
 $I(\cdot)$                       &  indicator function such that $I(\textrm{true}) = 1$ and $I(\textrm{false}) = 0$ \\ \hline
\end{tabular}
\end{center}
}
\end{table}

Table~\ref{tab:notation} summarizes the notation and some of the expressions used in this study.
The load $l_i$ is the amount of energy used during time interval $i$,
and $g_i$ is the amount of energy generated over the same period.
The residual energy in the battery at the end of interval $i$ is $x_i$.
We set the length of a time interval to one hour.
The energy supplied to the battery during time interval $i$ is $x_i - x_{i-1}$ and
the net energy drawn from the grid is $x_{i} - x_{i - 1} + l_{i} - g_{i}$.
Thus, the cost of energy over the time interval $i$ is $(x_{i} - x_{i - 1} + l_{i} - g_{i}) p_i$,
where $p_i$ is the price set for that interval.
The possibility of compensation tariff for feed-in electricity
is not considered in this study.
If such a tariff is high, scheduling will favor feeding electricity into the grid.
However, the trend in smart grid pricing is to encourage
residential users to conserve any electricity that they generate,
so feed-in tariffs are likely to become very low or zero, which
is what we assume.
In this case, the total cost of energy over $T$ time intervals is
$\sum_{i=1}^{T} I(x_{i} - x_{i - 1} + l_{i} - g_{i} > 0) (x_{i} - x_{i - 1} + l_{i} - g_{i}) p_i$,
where $I$ is the indicator function.
We use twenty-four hour data, and set $T$ to $24$.

The total cost of electricity is the sum of the energy charge and the demand charge,
which is the product of the fixed rate $p^*$ and the peak demand, and can thus be 
written $\max_{1 \le i \le T} \{ x_{i} - x_{i - 1} + l_{i} - g_{i} \} p^*$.
The problem of minimizing the total cost of electricity can now be expressed as follows:

Minimize
\begin{eqnarray*}
  & &   \sum_{i=1}^{T} I(x_{i} - x_{i - 1} + l_{i} - g_{i} > 0) (x_{i} - x_{i - 1} + l_{i} - g_{i}) p_i\\
  & &   ~~~~~~~~~                       + \max_{1 \le i \le T} \{ x_{i} - x_{i - 1} + l_{i} - g_{i} \} p^* ,\\
\end{eqnarray*}

subject to
\begin{eqnarray*}
  & &  0 \le x_i \le C, ~~~i = 1, 2, \ldots, T,\\
  & &  -P_d \le x_i - x_{i-1} \le P_c, ~~~i = 1, 2, \ldots, T,\\
\end{eqnarray*}
where $C$ is the total battery capacity,
$P_d$ is the battery discharge power, and $P_c$ is the battery charge power.
The value of $x_i$ cannot exceed the battery capacity,
and the net amount of energy $x_i - x_{i-1}$ flowing in or out of the battery
should lie in the range $[-P_d, P_c]$.

\section{Real-coded Genetic Algorithm}
\label{sec:rcga}


GAs that are based on real number representation are called real-coded
GAs (RCGAs) \cite{Herrera98}. Real coding was first used in specific applications,
such as chemometric problems and
in using meta-operators to find the most appropriate
parameters for a standard GA \cite{HerrLV98}. Subsequently, RCGAs have mainly been used
in numerical optimization problems over continuous domains \cite{Kim08, Yoon13-2, Yoon13, Yoon12}.



In our RCGA, a population consisting of 
$N/2$ pairs are randomly selected from a population of $N$,
and crossover and mutation operators are applied to each pair to generate $N/2$ offspring.
Both parents and offspring are ranked and the best $N$ become in the next generation.
We use a population of 100, and our RCGA terminates after 2,000 generations.

\subsection{Encoding}

Our RCGA is encoded using an array of $T$ real numbers.
Our approach differs from 
a typical real encoding in that
each gene $x_i$ has its own range of real values that are determined
by the value of its left-sided gene $x_{i-1}$.
The following two constraints must be satisfied by each value of $x_i$:

\begin{eqnarray*}
  & &  0 \le x_i \le C,\\
  & &  -P_d \le x_i - x_{i-1} \le P_c \Leftrightarrow x_{i - 1} - P_d \le x_i \le x_{i - 1} + P_c. \\
\end{eqnarray*}
Therefore $x_i$ must satisfy the following expression in $x_{i-1}$:

\begin{equation}
\label{eqn:gene}
\max (0, x_{i - 1} - P_d) \le x_i \le \min (C, x_{i - 1} + P_c)
\end{equation}

\subsection{Evaluation}

The objective function for the problem is used as 
the evaluation function of the RCGA:
$\sum_{i=1}^{T} I(y_{i} - y_{i - 1} + l_{i} - g_{i} > 0) (y_{i} - y_{i - 1} + l_{i} - g_{i}) p_i
+ \max_{1 \le i \le T} \{ y_{i} - y_{i - 1} + l_{i} - g_{i} \} p^*$.
Because this is a minimization problem,
solutions with smaller objective values are more likely to survive.

\subsection{Initialization}

Initially 100 solutions are generated, satisfying the feasibility constraint of Equation (\ref{eqn:gene}).
For each $i$ ($i = 1, 2, \ldots, T$), a real number is randomly chosen over the interval
$[\max (0, x_{i - 1} - P_d), \min (C, x_{i - 1} + P_c)]$.
Each solution generated by this procedure corresponds to an available ESS schedule.

\subsection{Crossover Operator}

We use the crossover operator BLX-$\alpha$ \cite{Brem66, Eshe93},
where $\alpha$ is a nonnegative real-valued parameter.
This operator produces $z_k = (z_1, z_2, \ldots, z_n)$ offspring,
where $z_i$ is a random number chosen over the interval
$[C_{\min} - \alpha I, C_{\max} + \alpha I]$, where $C_{\max} = \max(x_i, y_i)$,
$C_{\min} = \min(x_i, y_i)$, and $I = C_{\max} - C_{\min}$. 
The value of $\alpha$ is set to $0.5$ in our RCGA.
To ensure that each gene satisfies Equation (\ref{eqn:gene}),
BLX-$\alpha$ is modified
so that it accepts
random real numbers over the interval
$[\max (0, x_{i - 1} - P_d, C_{\min} - \alpha I), \min (C, x_{i - 1} + P_c), C_{\max} + \alpha I)]$,
instead of the interval $[C_{\min} - \alpha I, C_{\max} + \alpha I]$.

\subsection{Mutation Operator}

In Gaussian mutation \cite{GoldGen89}, the $i^{\textrm th}$ parameter $x_i$ of an
individual is mutated by $x_i = x_i + N(0, \sigma_i)$
at a mutation rate $p_m$, where $N(0, \sigma_i)$
is an independent random Gaussian number with a mean of zero
and a standard deviation of $\sigma_i$. In our RCGA,
$\sigma_i$ is set to $\min (C, x_{i - 1} + P_c) - \max (0, x_{i - 1} - P_d)$, which is 
 the magnitude of the range of feasible solutions.
If a mutated value is not in $[\min (C, x_{i - 1} + P_c), \max (0, x_{i - 1} - P_d)]$,
then it is replaced by
$\min (C, x_{i - 1} + P_c)$ or $\max (0, x_{i - 1} - P_d)$, whichever is closest,
to produce a feasible solution.
If $i < j \le T$,
then changes to $x_i$ can affect the feasibility of $x_j$.
Thus, 
values of $x_j$ which are not in $[\min (C, x_{j - 1} + P_c), \max (0, x_{j - 1} - P_d)]$,
are similarly replaced by the closer of 
$\min (C, x_{j - 1} + P_c)$ or $\max (0, x_{j - 1} - P_d)$.
The value of $p_m$ is set to $0.1/T$.

\section{Simulation Results}
\label{sec:results}

\subsection{Problems Instances}

\begin{table}[tb!]
\caption{Residential customer load profile data used in this study.\label{tab:load}}
{\small
\begin{center}
\begin{tabular}{|l|l|} \hline
Seasons & summer (Jun. -- Sep.) \\ 
        & winter (Dec. -- Feb.) \\ \hline 
Types of day & weekday \\
             & weekend \\ \hline
Weather scenarios & normal \\ \hline
\end{tabular}
\end{center}
}
\end{table}

\begin{table}[tb!]
\caption{PV system specifications.\label{tab:pv}}
{\small
\begin{center}
\begin{tabular}{l|l} \hline
DC rating  & 3 kW \\
DC to AC derating factor & 0.77 \\
Array type   & fixed tilt \\
Array tilt   & $46.6^{\circ}$ (latitude) \\
Array azimuth   & $180.0^{\circ}$ (true south) \\ \hline
\end{tabular}
\end{center}
}
\end{table}

\begin{table}[tb!]
\caption{Time-of-use prices used in this study (USD).\label{tab:tou}}
{\small
\begin{center}
\begin{tabular}{|c|c|c|} \hline
Hour (from | to) & Summer (cents/kWh) & Winter (cents/kWh) \\\hline \hline
0 -- 1   & \cellcolor{yellow} 5 & \cellcolor{yellow} 5 \\ \hline
1 -- 2   &\cellcolor{yellow}  5 & \cellcolor{yellow} 5 \\ \hline
2 -- 3   &\cellcolor{yellow}  5  &\cellcolor{yellow}  5 \\ \hline
3 -- 4   &\cellcolor{yellow}  5  &\cellcolor{yellow}  5 \\ \hline
4 -- 5   &\cellcolor{yellow}  5  &\cellcolor{yellow}  5 \\ \hline
5 -- 6   &\cellcolor{yellow}  5  &\cellcolor{yellow}  5 \\ \hline
6 -- 7   &\cellcolor{yellow}  5  &\cellcolor{yellow}  5 \\ \hline
7 -- 8   &  \cellcolor{orange} 10   & \cellcolor{red} 15  \\ \hline
8 -- 9   & \cellcolor{orange} 10     & \cellcolor{red} 15 \\ \hline
9 -- 10  & \cellcolor{orange} 10     & \cellcolor{red} 15 \\ \hline
10 -- 11 & \cellcolor{orange} 10 & \cellcolor{red} 15 \\ \hline
11 -- 12  &    \cellcolor{red} 15   &  \cellcolor{orange} 10 \\ \hline
12 -- 13  &  \cellcolor{red} 15   &  \cellcolor{orange} 10 \\ \hline
13 -- 14  &  \cellcolor{red} 15   &  \cellcolor{orange} 10 \\ \hline
14 -- 15  &  \cellcolor{red} 15   &  \cellcolor{orange} 10 \\ \hline
15 -- 16  &  \cellcolor{red} 15   &  \cellcolor{orange} 10 \\ \hline
16 -- 17  &  \cellcolor{red} 15   &  \cellcolor{orange} 10 \\ \hline
17 -- 18&    \cellcolor{orange} 10 &     \cellcolor{red} 15 \\ \hline
18 -- 19 &   \cellcolor{orange} 10  &    \cellcolor{red} 15 \\ \hline
19 -- 20 &\cellcolor{yellow} 5  &\cellcolor{yellow}  5  \\ \hline
20 -- 21 &\cellcolor{yellow} 5  &\cellcolor{yellow}  5  \\ \hline
21 -- 22&\cellcolor{yellow}  5  &\cellcolor{yellow}  5  \\ \hline
22 -- 23&\cellcolor{yellow}  5  &\cellcolor{yellow}  5  \\ \hline
23 -- 24&\cellcolor{yellow}  5  &\cellcolor{yellow}  5  \\ \hline \hline
$p^*$ & \multicolumn{2}{|c|}{Demand charge rate: 20 (low), 30 (high) (cents/kW)} \\ \hline
\end{tabular}
\end{center}
}
\end{table}

The load profile that we will use is
a residential customer profile provided by NorthWestern Energy
\cite{NWloadprofile},
and is based on data for 1992 and 1993.
The company's 
Load Vision profiling software was used to construct
profiles for typical diversified residential loads
on weekdays and at weekends, for
each season and three weather scenarios.
The portion of the data that we used in this study is presented as Table \ref{tab:load}.

Hourly solar generation data were obtained using PVWatts,
developed by the National Renewable Energy Laboratory (NREL).
This calculator predicts the energy production of
residential and small commercial PV installations,
based on hourly data for sunny and cloudy days in
Helena, a city in the northwestern United States.
The specifications of the PV system that we consider are listed in Table \ref{tab:pv}.

Typical TOU prices were generated by simulations using three price levels, for summer and winter,
based on the TOU pricing models of several utility companies.
We consider two daily rates of demand charge,
20 cents/kW (low) and 30 cents/kW (high).
The TOU pricing model that we have constructed is given in Table \ref{tab:tou}.

We consider a battery with a total capacity of 2 kWh,
but only 1.8 kWh is used to extend battery life.
The maximum rate of charge and discharge is around to be 0.6 kW.
Thus, we set $C$ to 1.8, and
and values of $P_c$ and $P_d$ of 0.6 kW are used in the problem formulation.

\subsection{Results and Discussion}

We compared our RCGA with a net-power-based algorithm (NPB), which
charges or discharges the battery to make up the difference between the power generated and the load.
This naive algorithm does not consider the electricity price at all.

\begin{table}[tb!]
\caption{Comparison of simulation results for a single day.\label{tab:results_noeff}}
{\small
\begin{center}
\begin{tabular}{c|l|l|l|l||r||r|r||r|r} \hline
\multicolumn{5}{c||}{Instance}  & NO-ESS    & \multicolumn{2}{c||}{NPB} & \multicolumn{2}{c}{RCGA} \\ \hline
Case & Rate& Season  & Weather & Day type &   Cost    &   Cost    &   Saving    &   Ave. cost (Std) & Saving \\ \hline \hline
1& Low  &   Summer  &   Sunny   &   Weekday &   83.69   &   68.76   &   18  &   63.87   (0.63)  &   24  \\
2& Low  &   Summer  &   Sunny   &   Weekend &   82.28   &   65.42   &   20  &   62.90   (0.61)  &   24  \\
3& Low  &   Summer  &   Cloudy  &   Weekday &   104.64  &   91.42   &   13  &   86.19   (1.09)  &   18  \\
4& Low  &   Summer  &   Cloudy  &   Weekend &   111.40  &   107.71  &   3   &   98.32   (1.09)  &   12  \\
5& Low  &   Winter  &   Sunny   &   Weekday &   185.43  &   161.05  &   13  &   150.08  (0.97)  &   19  \\
6& Low  &   Winter  &   Sunny   &   Weekend &   176.93  &   152.00  &   14  &   140.67  (1.22)  &   20  \\
7& Low  &   Winter  &   Cloudy  &   Weekday &   225.68  &   221.86  &   2   &   197.37  (0.74)  &   13  \\
8& Low  &   Winter  &   Cloudy  &   Weekend &   233.26  &   233.26  &   0   &   207.44  (0.69)  &   11  \\
9& High &   Summer  &   Sunny   &   Weekday &   94.37   &   79.44   &   16  &   71.75   (0.72)  &   24  \\
10& High    &   Summer  &   Sunny   &   Weekend &   92.47   &   75.44   &   18  &   70.43   (0.76)  &   24  \\
11& High    &   Summer  &   Cloudy  &   Weekday &   115.32  &   102.10  &   11  &   96.32   (1.20)  &   16  \\
12& High    &   Summer  &   Cloudy  &   Weekend &   121.42  &   117.73  &   3   &   108.42  (1.31)  &   11  \\
13& High    &   Winter  &   Sunny   &   Weekday &   201.45  &   176.88  &   12  &   163.75  (1.26)  &   19  \\
14& High    &   Winter  &   Sunny   &   Weekend &   192.56  &   166.84  &   13  &   153.75  (1.33)  &   20  \\
15& High    &   Winter  &   Cloudy  &   Weekday &   241.70  &   237.88  &   2   &   211.49  (1.09)  &   12  \\
16& High    &   Winter  &   Cloudy  &   Weekend &   248.89  &   248.89  &   0   &   221.26  (0.90)  &   11  \\ \hline
\end{tabular}
\end{center}
\hspace*{0.0cm} NO-ESS is the cost with no ESS.\\
\hspace*{0.0cm} The NPB algorithm charges the battery when the generated power exceeds the load and discharges otherwise.\\
\hspace*{0.0cm} RCGA is our real-coded genetic algorithm (the average costs are obtained over 100 runs). \\
\hspace*{0.0cm} All costs are in US cents, and savings are percentages. \\
\hspace*{0.0cm} The saving for Algorithm $A$ is obtained using the formula, $100 \times (Cost_{\textrm{\scriptsize NO-ESS}} - Cost_A) / Cost_{\textrm{\scriptsize NO-ESS}}$, where $Cost_A$ is the electricity cost incurred by Algorithm $A$.
\\
}
\end{table}

The simulation results are shown in Table \ref{tab:results_noeff}.
All the algorithms run in under a second on an Intel Xeon CPU E5530 @ 2.40 GHz.
A run of RCGA takes 0.18 seconds.

The results in Table \ref{tab:results_noeff} show
that our RCGA always outperforms the NPB.
The maximum benefit is $11\%$ in Cases 7, 8, and 16, and
the minimum is $4\%$ in Case 2.
The RCGA performed better in the winter than in the summer.
This result can be explained by the difference in the summer and winter price schedules.
In summer, the peak period usually occurs
at a time when PV energy is plentiful,
and so the power drawn from the grid is easily reduced,
without the need for an elaborate algorithm.
In winter, there is much less PV energy available during peak periods,
making the RCGA more effective.

\begin{figure*}[tb!]
\centerline{
\begin{minipage}[tb]{3.3in}
\psfig{figure=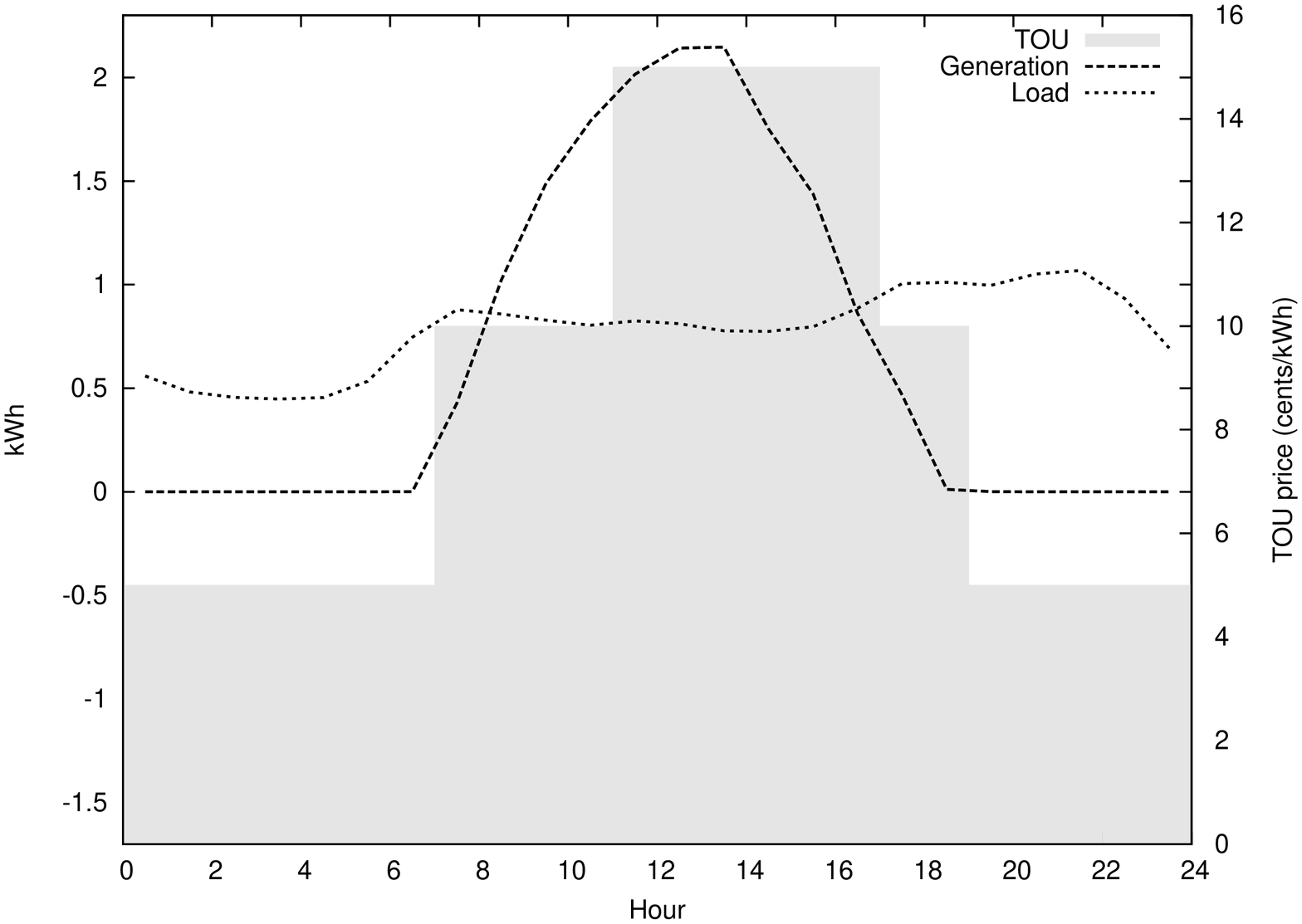,width=2.25in,angle=-90}\\
{\scriptsize (a) Sunny day (Case 1): PV and load profiles with TOU}
\end{minipage}
\begin{minipage}[tb]{3.3in}
\psfig{figure=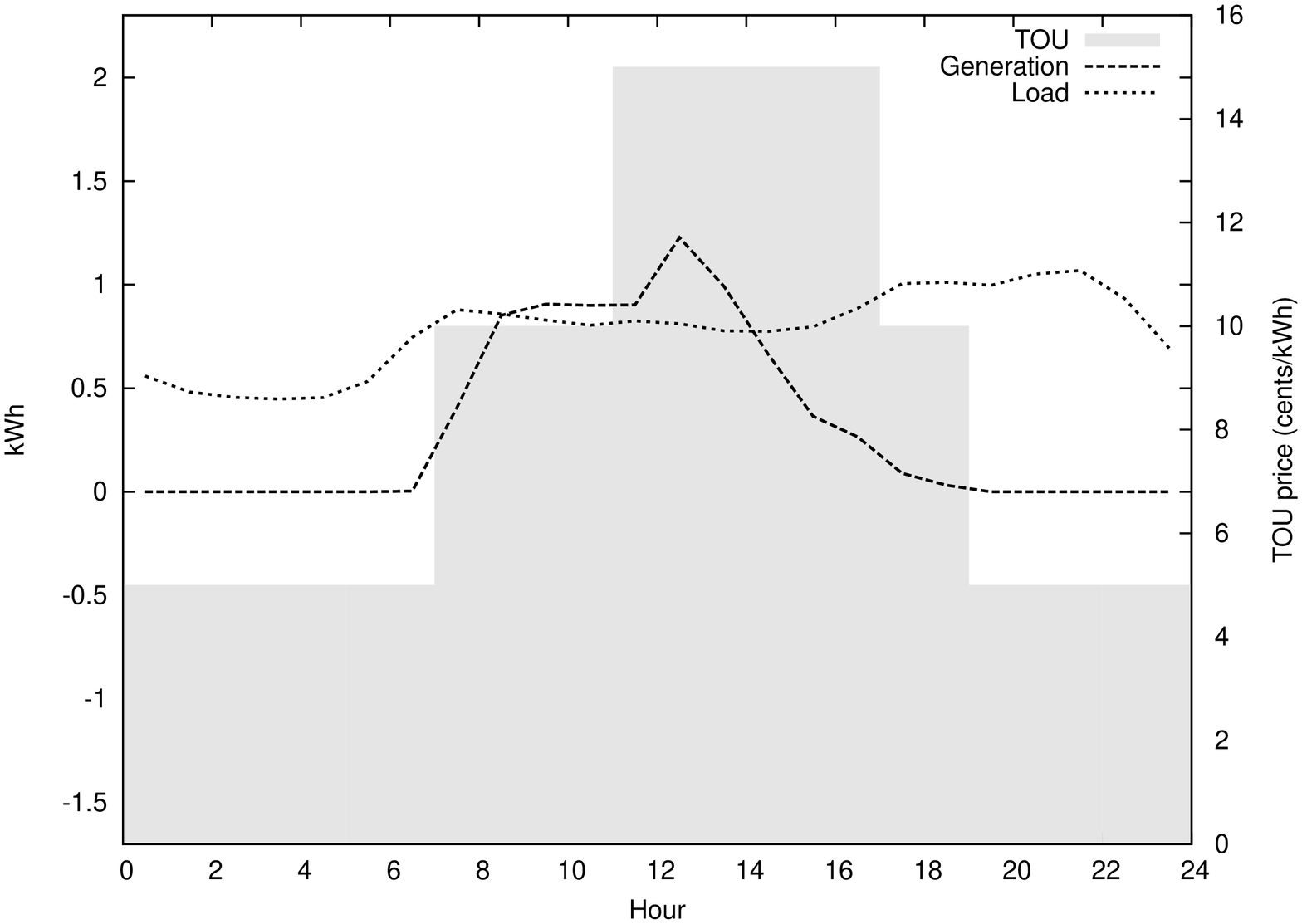,width=2.25in,angle=-90}\\
{\scriptsize (d) Cloudy day (Case 3): PV and load profiles with TOU}
\end{minipage}
}
\centerline{
\begin{minipage}[tb]{3.3in}
\psfig{figure=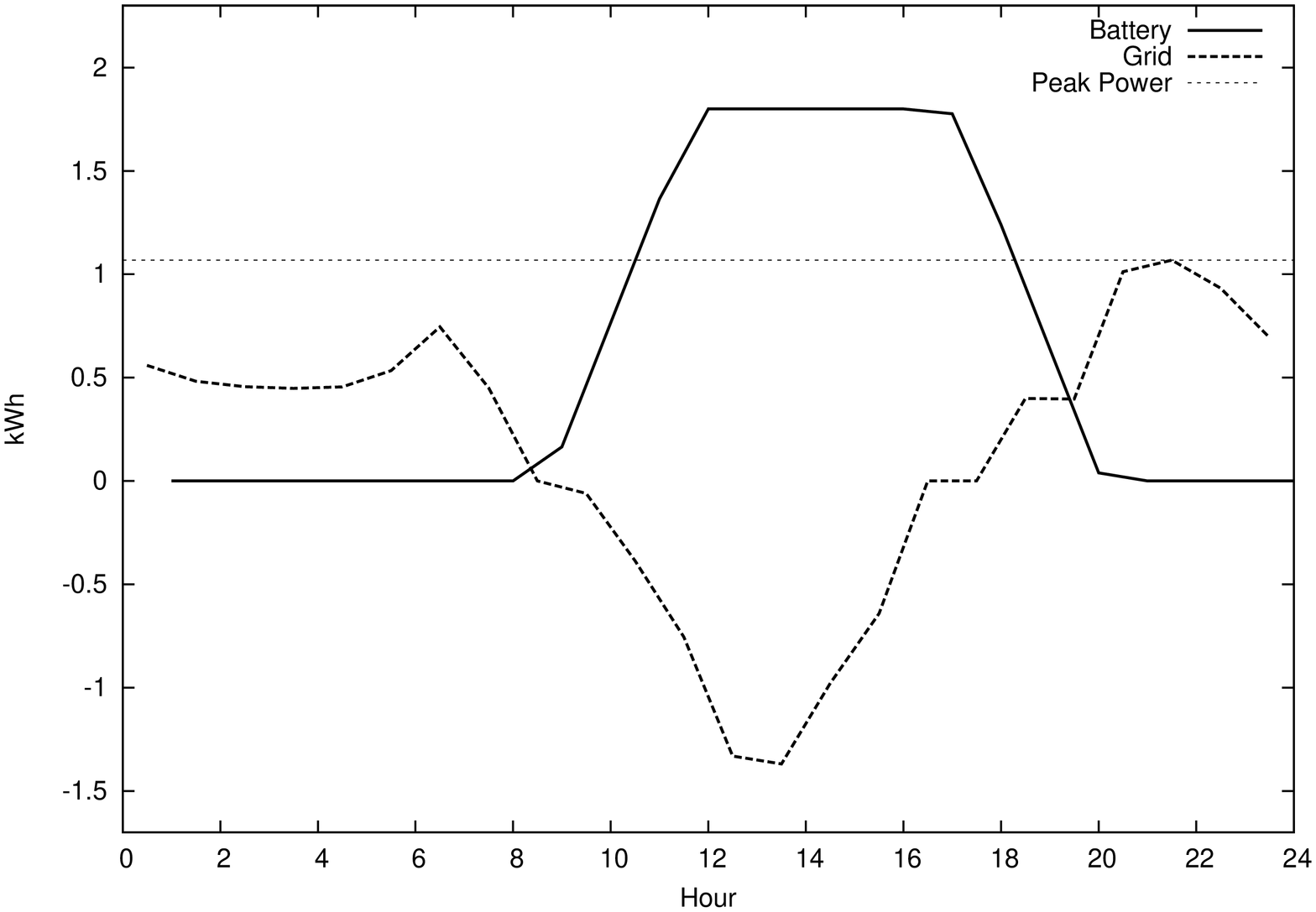,width=2.07in,angle=-90}\\
{\scriptsize (b) Sunny day (Case 1): ESS scheduled by NPB}
\end{minipage}
\begin{minipage}[tb]{3.3in}
\psfig{figure=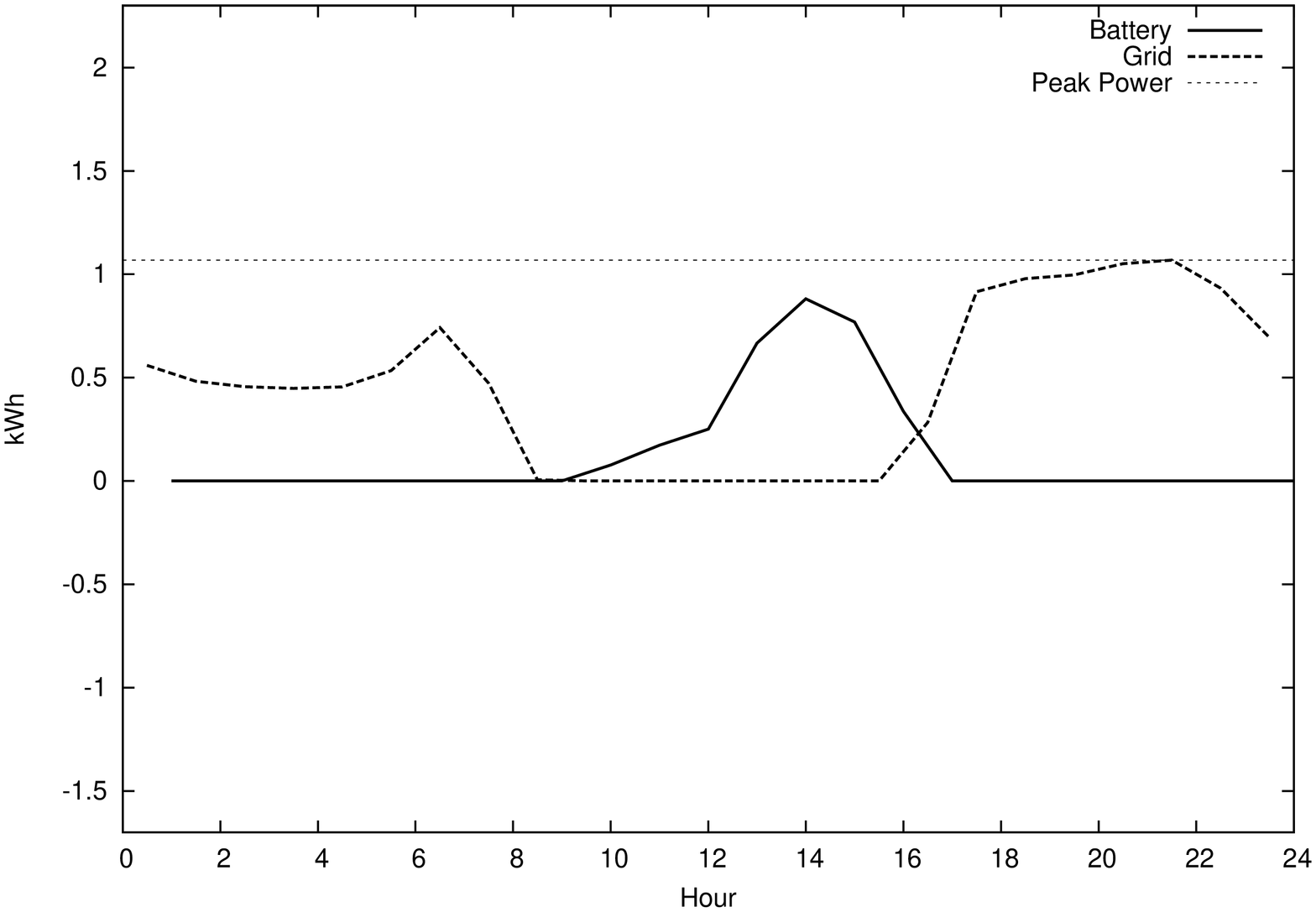,width=2.07in,angle=-90}\\
{\scriptsize (e) Cloudy day (Case 3): ESS scheduled by NPB}
\end{minipage}
}
\centerline{
\begin{minipage}[tb]{3.3in}
\psfig{figure=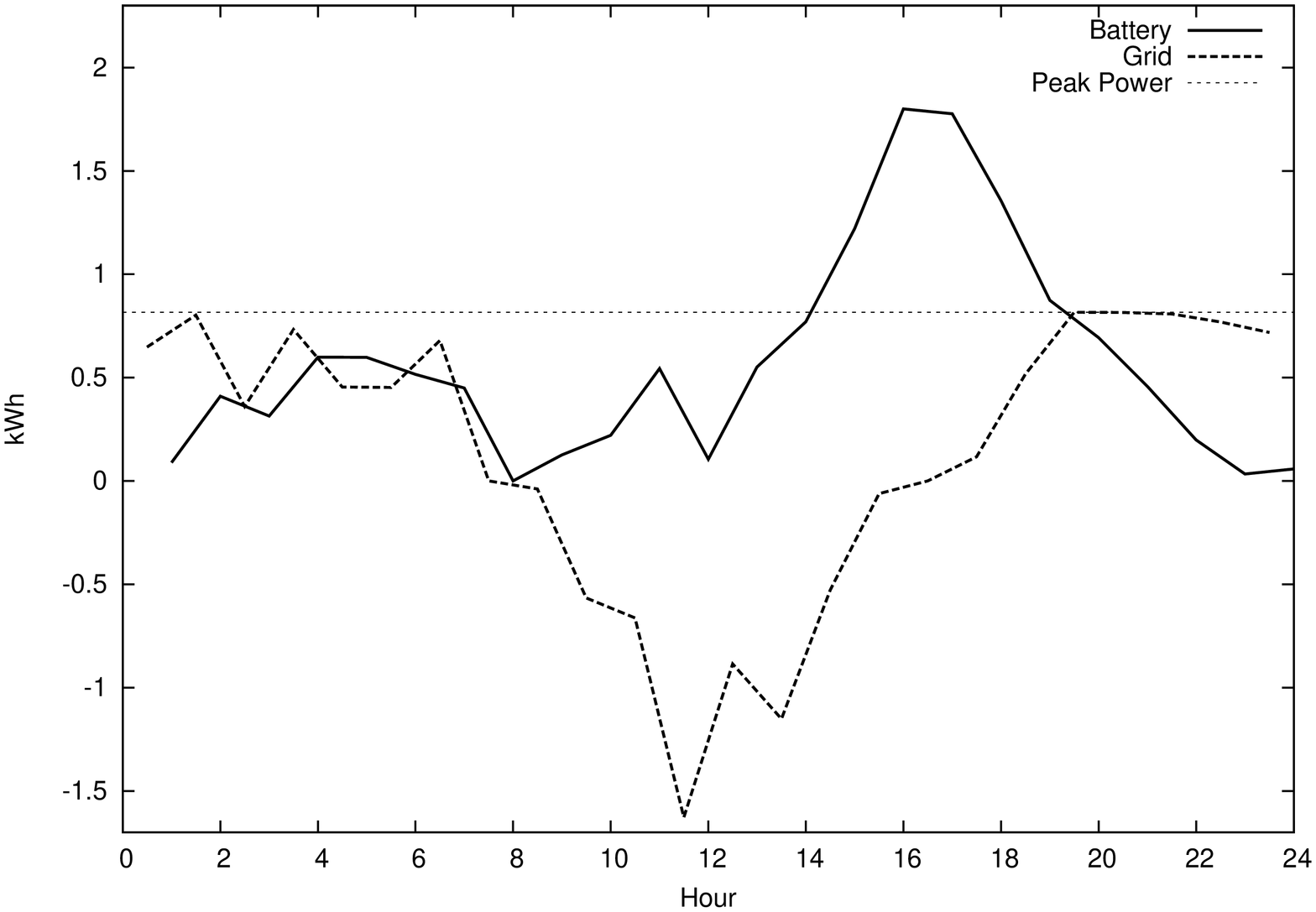,width=2.07in,angle=-90}\\
{\scriptsize (c) Sunny day (Case 1): ESS scheduled by RCGA}
\end{minipage}
\begin{minipage}[tb]{3.3in}
\psfig{figure=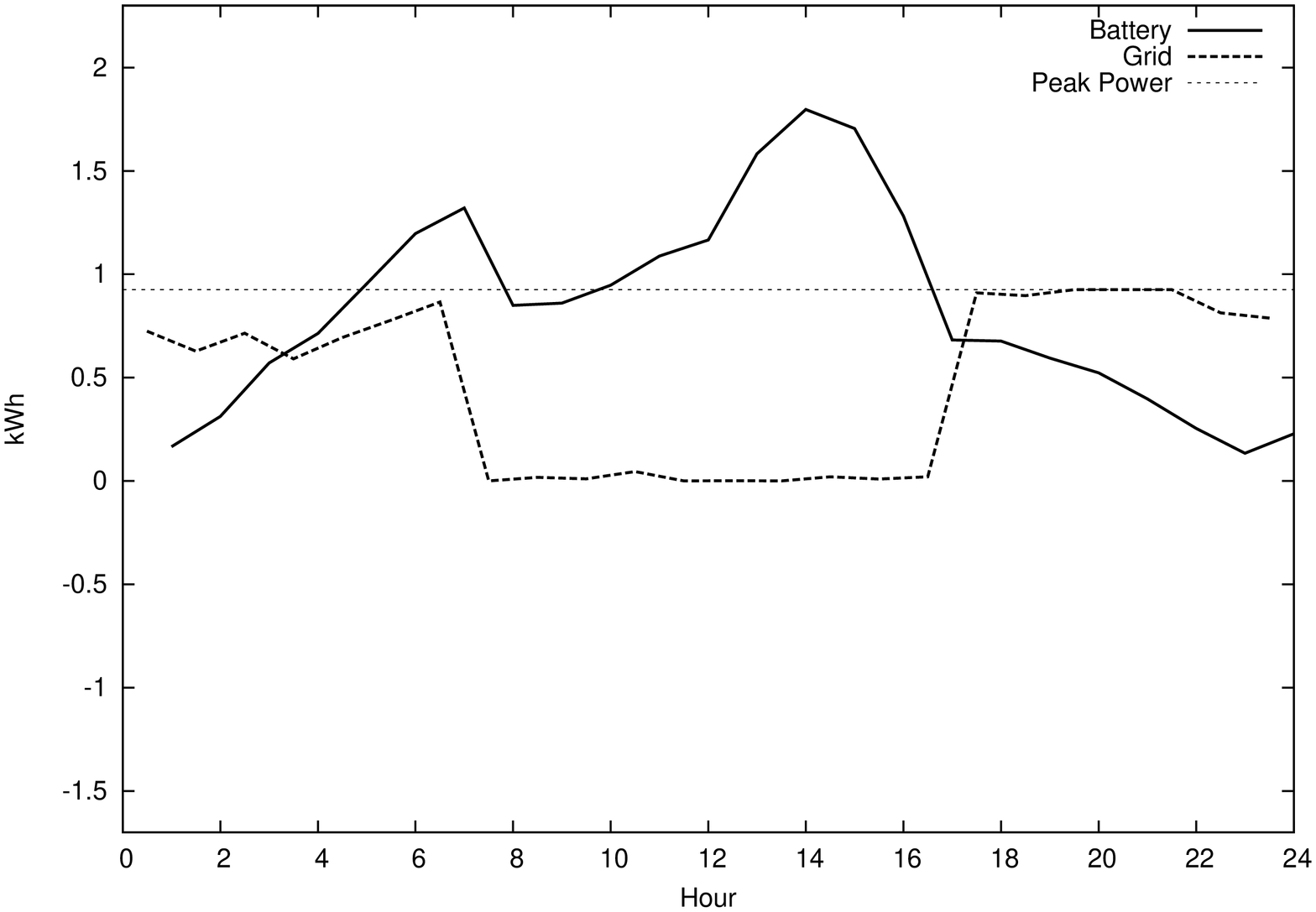,width=2.07in,angle=-90}\\
{\scriptsize (f) Cloudy day (Case 3): ESS scheduled by RCGA}
\end{minipage}
}
\caption[]{Simulated battery schedules for summer weekdays.}
\label{fig:summer}
\end{figure*}
\begin{figure*}[tb!]
\centerline{
\begin{minipage}[tb]{3.3in}
\psfig{figure=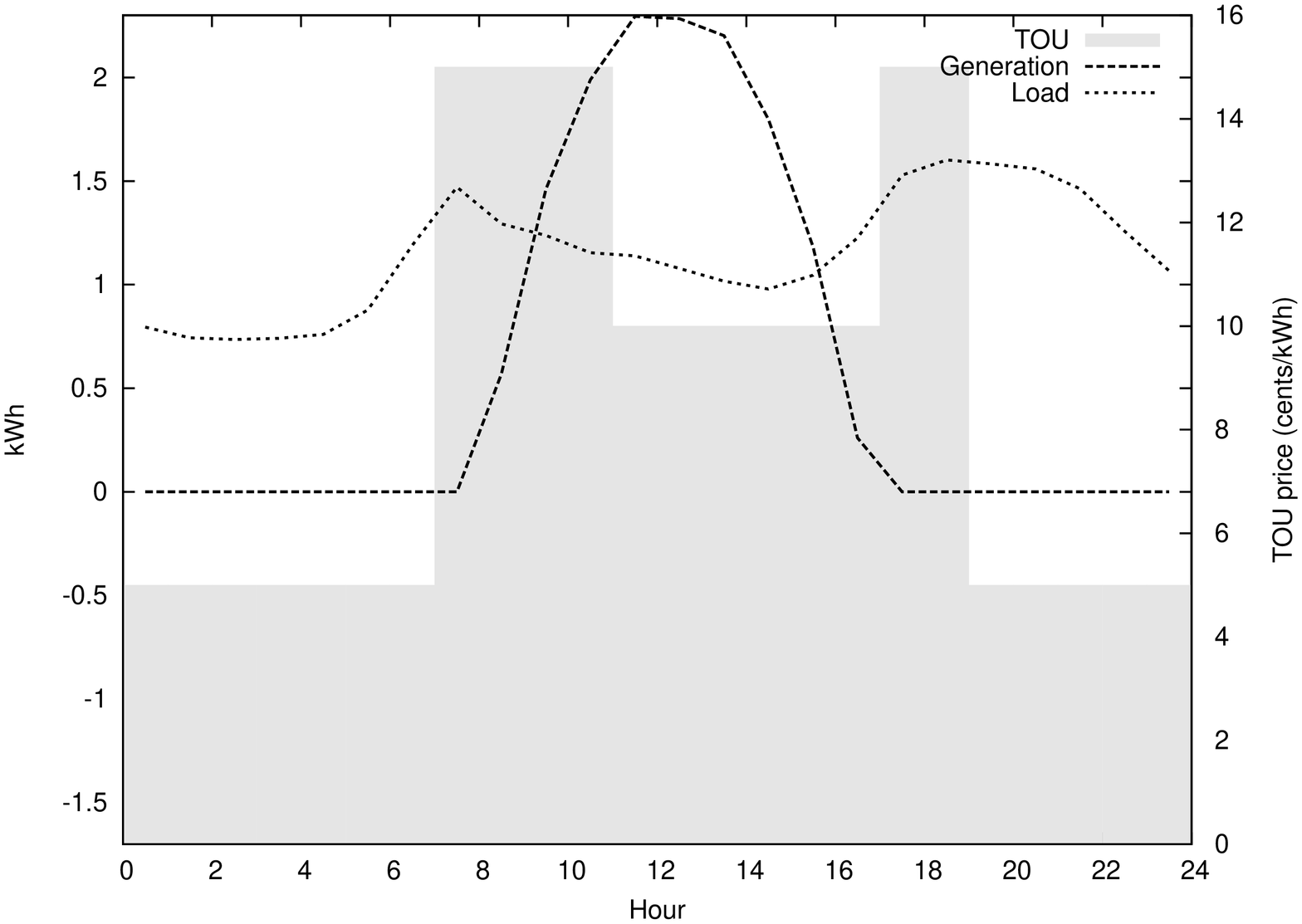,width=2.25in,angle=-90}\\
{\scriptsize (a) Sunny day (Case 5): PV and load profiles with TOU}
\end{minipage}
\begin{minipage}[tb]{3.3in}
\psfig{figure=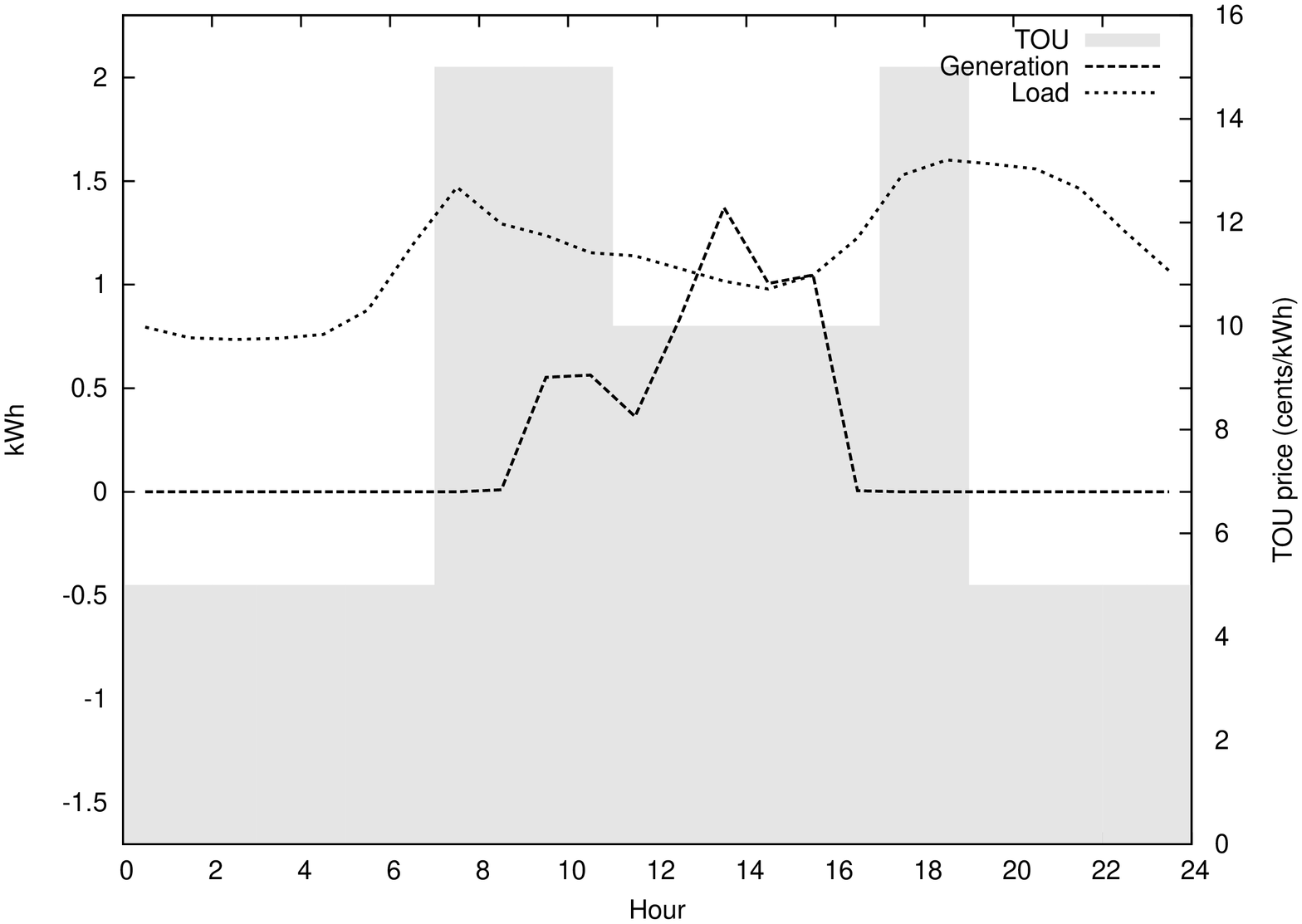,width=2.25in,angle=-90}\\
{\scriptsize (d) Cloudy day (Case 7): PV and load profiles with TOU}
\end{minipage}
}
\centerline{
\begin{minipage}[tb]{3.3in}
\psfig{figure=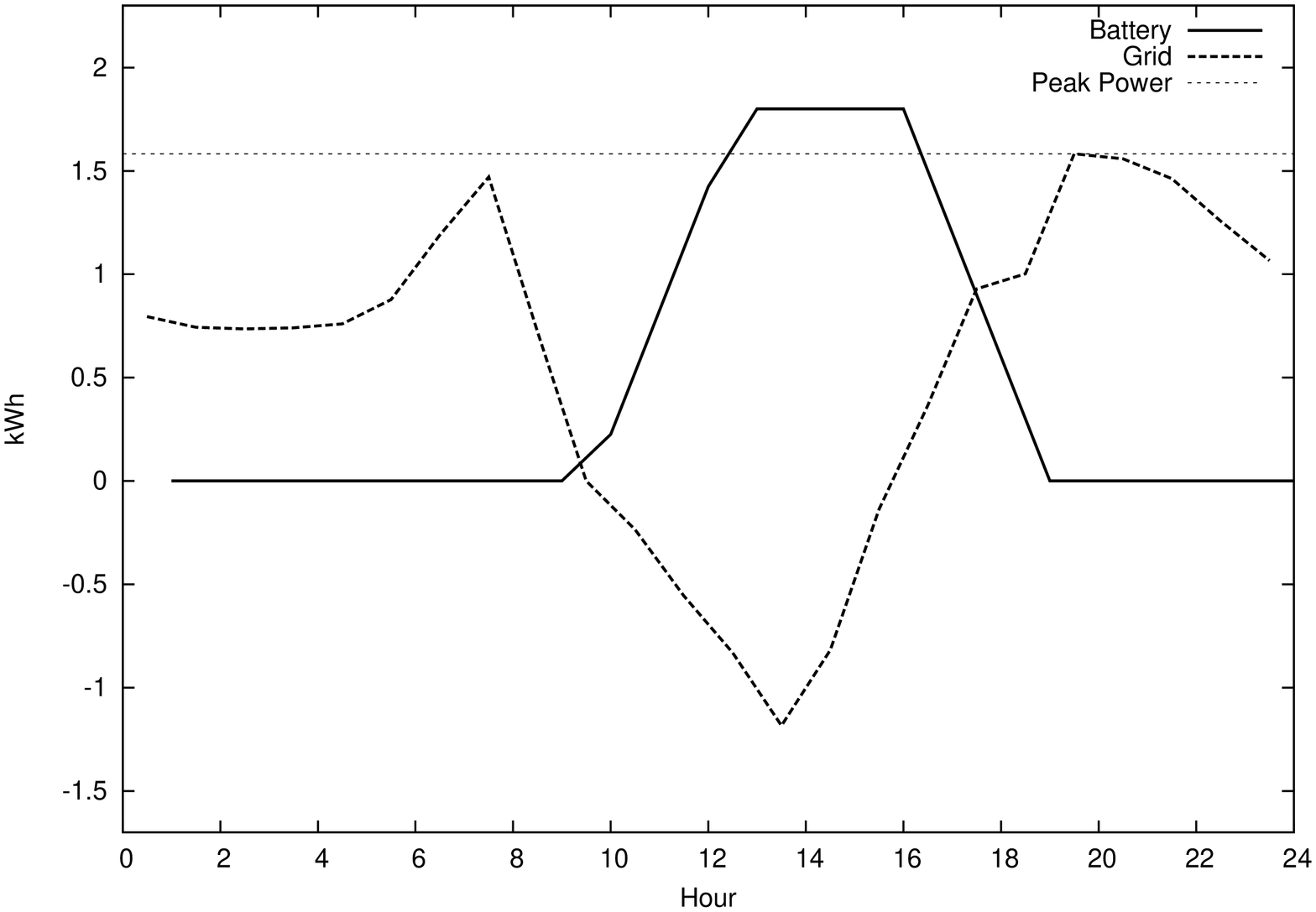,width=2.07in,angle=-90}\\
{\scriptsize (b) Sunny day (Case 5): ESS scheduled by NPB}
\end{minipage}
\begin{minipage}[tb]{3.3in}
\psfig{figure=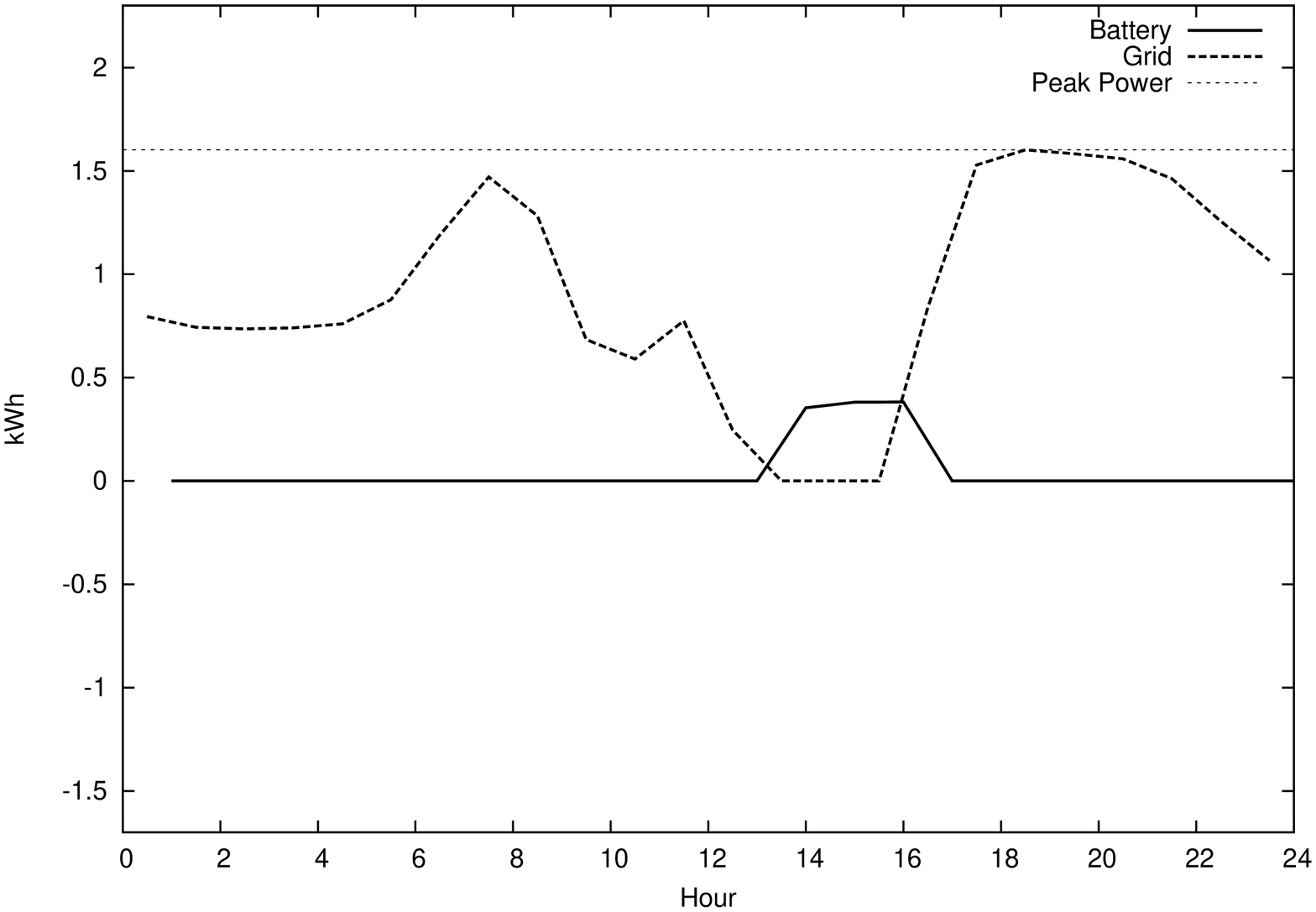,width=2.07in,angle=-90}\\
{\scriptsize (e) Cloudy day (Case 7): ESS scheduled by NPB}
\end{minipage}
}
\centerline{
\begin{minipage}[tb]{3.3in}
\psfig{figure=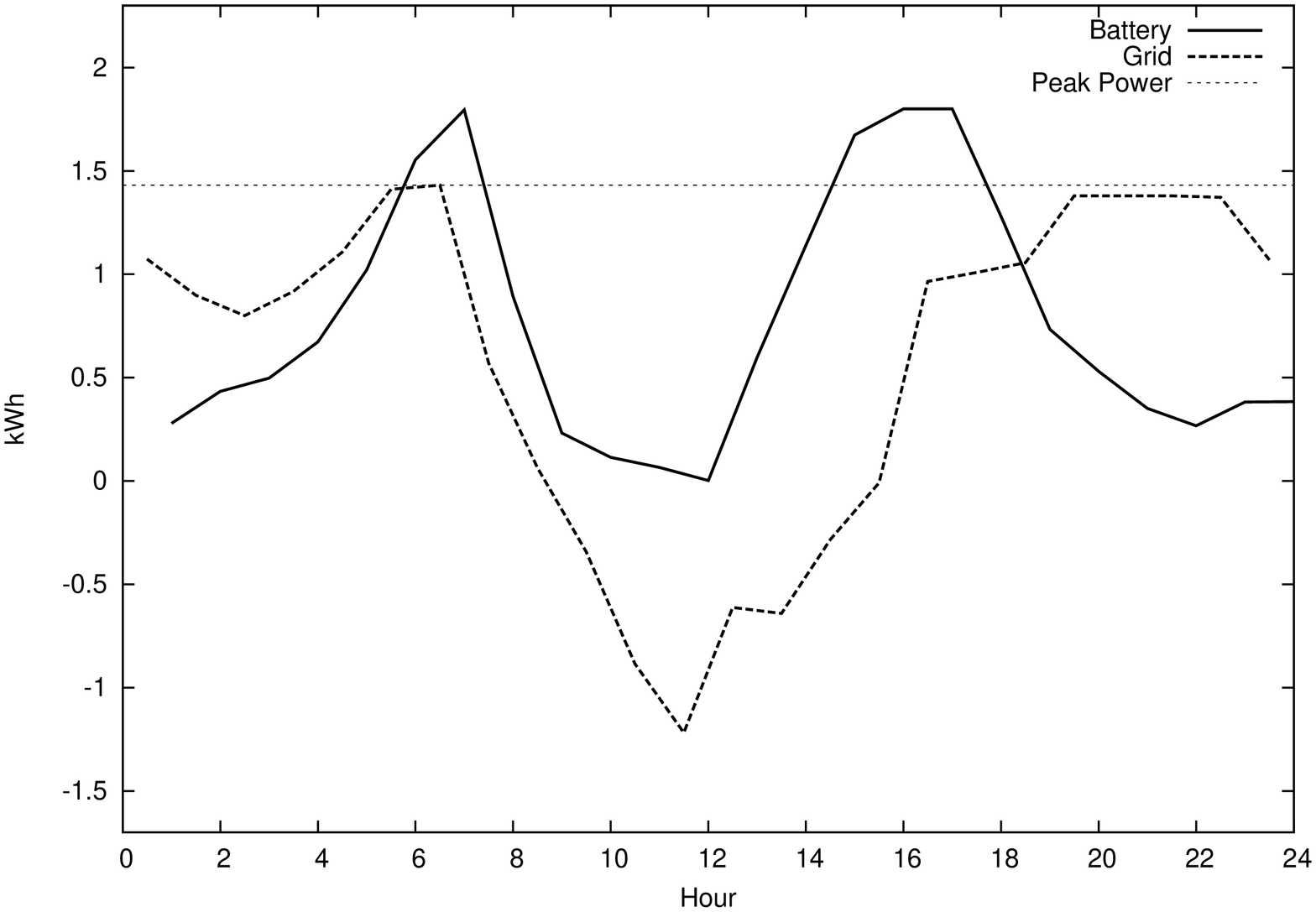,width=2.07in,angle=-90}\\
{\scriptsize (c) Sunny day (Case 5): ESS scheduled by RCGA}
\end{minipage}
\begin{minipage}[tb]{3.3in}
\psfig{figure=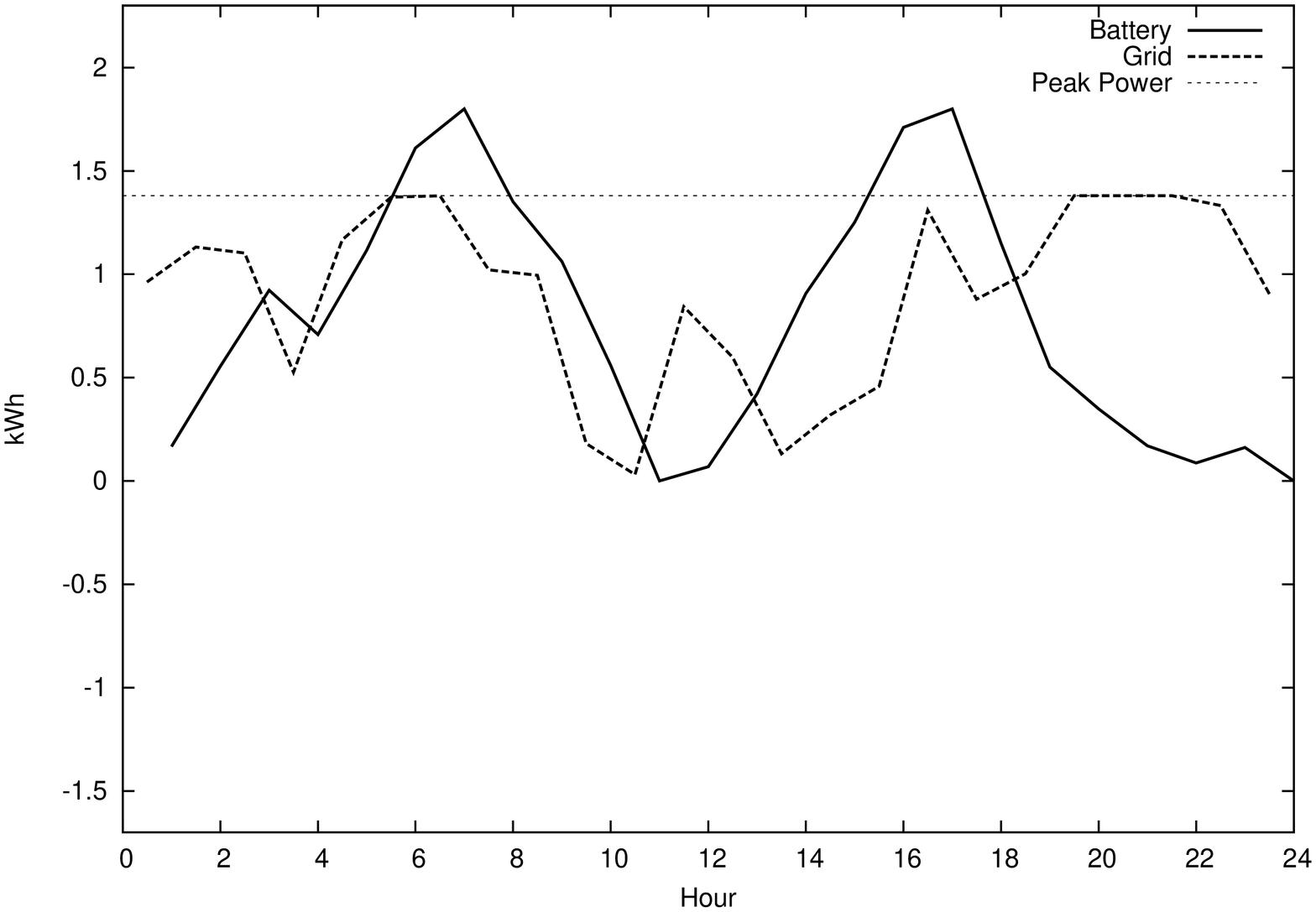,width=2.07in,angle=-90}\\
{\scriptsize (f) Cloudy day (Case 7): ESS scheduled by RCGA}
\end{minipage}
}
\caption[]{Simulated battery schedules for winter weekdays}
\label{fig:winter}
\end{figure*}

Figs.~\ref{fig:summer} and \ref{fig:winter} show simulated levels of battery charge
which are produced by the NPB and RCGA in the winter and the summer scenarios.
Typical PV and load profiles with TOU prices for each season and weather scenario are shown
in Figs.~\ref{fig:summer}(a), \ref{fig:summer}(d), \ref{fig:winter}(a),
and \ref{fig:winter}(d).
The other figures show battery charge, the average and peak power drawn from the grid.
The NPB charges the battery when the generation exceeds the load and discharges otherwise.
The ESS operated on the NPB schedule charges the battery in the daytime when the sun blazes
and discharges the battery in the evening when the sun sets, regardless of the season and the weather.
However, the RCGA optimizes the ESS charge schedule to minimize the electricity cost
under various constraints and a given pricing policy.
Thus, the ESS operated on the RCGA schedule charges and discharges the battery dynamically depending on the season and the weather.
The figures show that the RCGA schedules reduce both the peak power and the
purchase of electricity (i.e., grid power) during on-peak periods.

\section{Conclusion}
\label{sec:conclusion}

We have developed an RCGA for ESS charge scheduling, which
is especially important for electricity customers who have to contend with dynamic pricing.
We considered TOU pricing with a demand charge, 
when electricity is supplied to a customer with their own renewable energy generation facility.
The scheduling problem for this scenario was formally defined,
and the RCGA was used to develop a novel approach to charge scheduling.
Experiments using the load and generation profiles of typical residential customers showed
that scheduling by the RCGA
reduced both the peak power consumption and the purchase of electricity during on-peak periods.
This suggests that charge scheduling using an RCGA can help to reduce customers' electricity bills.

Neither battery efficiency nor the capital cost of a storage system
were considered in this study, although
these factors can clearly affect overall cost.
Further studies that consider these factors are needed.
It would also be interesting to investigate how our RCGA
performs under more dynamic pricing schemes such as real-time pricing,
which are a part of many smart grid scenarios.


\subsection*{Conflict of Interests}
The authors declare that there is no conflict of interests regarding
the publication of this paper.


\begin{thebibliography}{10}

\bibitem{NWloadprofile}
Northwestern {E}nergy -- {R}esidential {C}ustomer {P}rofile.
\newblock
  \\http://www.northwesternenergy.com/display.aspx?Page=Supplier\_Residential/.

\bibitem{barton2004energy}
J.~P. Barton and D.~G. Infield.
\newblock Energy storage and its use with intermittent renewable energy.
\newblock {\em IEEE Transactions on Energy Conversion}, 19(2):441--448, 2004.

\bibitem{Brem66}
H.~J. Bremermann, J.~Rogson, and S.~Salaff.
\newblock Global properties of evolution processes.
\newblock In H.~H. Pattee, editor, {\em Natural Automata and Useful
  Simulations}, pages 3--42, 1966.

\bibitem{cao2012optimized}
Y.~Cao, S.~Tang, C.~Li, P.~Zhang, Y.~Tan, Z.~Zhang, and J.~Li.
\newblock An optimized {EV} charging model considering {TOU} price and {SOC}
  curve.
\newblock {\em IEEE Transactions on Smart Grid}, 3(1):388--393, 2012.

\bibitem{Ven11}
P.~Van de~Ven, N.~Hegde, L.~Massouli{\'e}, and T.~Salonidis.
\newblock Optimal control of residential energy storage under price
  fluctuations.
\newblock In {\em Proceedings of the First International Conference on Smart
  Grids, Green Communications and IT Energy-aware Technologies}, pages
  159--162, 2011.

\bibitem{Eshe93}
L.~J. Eshelman and J.~D. Schaffer.
\newblock Real-coded genetic algorithms and interval-schemata.
\newblock In {\em Proceedings of the Second Workshop on Foundations of Genetic
  Algorithms}, pages 187--202, 1993.

\bibitem{Gallo13}
D.~Gallo, C.~Landi, M.~Luiso, and R.~Morello.
\newblock Optimization of experimental model parameter identification for
  energy storage systems.
\newblock {\em Energies}, 6(9):4572--4590, 2013.

\bibitem{GoldGen89}
D.~Goldberg.
\newblock {\em Genetic Algorithms in Search, Optimization, and Machine
  Learning}.
\newblock {Addison-Wesley}, 1989.

\bibitem{grillo2012optimal}
S.~Grillo, M.~Marinelli, S.~Massucco, and F.~Silvestro.
\newblock Optimal management strategy of a battery-based storage system to
  improve renewable energy integration in distribution networks.
\newblock {\em IEEE Transactions on Smart Grid}, 3(2):950--958, 2012.

\bibitem{Herrera98}
F.~Herrera, M.~Lozano, and J.~L. Verdegay.
\newblock Tackling real-coded genetic algorithms: operators and tools for
  behavioural analysis.
\newblock {\em Artificial Intelligence Review}, 12:265--319, 1998.

\bibitem{HerrLV98}
F.~Herrera, M.~Lozano, and J.~L. Verdegay.
\newblock Tackling real-coded genetic algorithms: Operators and tools for
  behavioural analysis.
\newblock {\em Artificial Intelligence Review}, 12(4):265--319, 1998.

\bibitem{Hu10}
W.~Hu, Z.~Chen, and B.~Bak-Jensen.
\newblock Optimal operation strategy of battery energy storage system to
  real-time electricity price in {Denmark}.
\newblock In {\em Proceedings of the IEEE Power and Energy Society General
  Meeting}, pages 1--7, July 2010.

\bibitem{Huang11}
T.~Huang and D.~Liu.
\newblock Residential energy system control and management using adaptive
  dynamic programming.
\newblock In {\em Proceedings of the International Joint Conference on Neural
  Networks}, pages 119--124, 2011.

\bibitem{hwangbo2013application}
S.-W. Hwangbo, B.-J. Kim, and J.-H. Kim.
\newblock Application of economic operation strategy on battery energy storage
  system at {Jeju}.
\newblock In {\em Proceedings of the IEEE PES Conference on Innovative Smart
  Grid Technologies Latin America (ISGT LA)}, pages 1--8. IEEE, 2013.

\bibitem{Kim08}
Y.-H. Kim, Y.~Yoon, and B.-R. Moon.
\newblock A {L}agrangian approach for multiple personalized campaigns.
\newblock {\em IEEE Transactions on Knowledge and Data Engineering},
  20(3):383--396, 2008.

\bibitem{Koutsopoulos11}
I.~Koutsopoulos, V.~Hatzi, and L.~Tassiulas.
\newblock Optimal energy storage control policies for the smart power grid.
\newblock In {\em Proceedings of the IEEE International Conference on Smart
  Grid Communications (SmartGridComm)}, pages 475 --480, Oct. 2011.

\bibitem{Lee07}
T.-Y. Lee.
\newblock Operating schedule of battery energy storage system in a time-of-use
  rate industrial user with wind turbine generators: a multipass iteration
  particle swarm optimization approach.
\newblock {\em IEEE Transactions on Energy Conversion}, 22(3):774--782, 2007.

\bibitem{lee1995determination}
T.-Y. Lee and N.~Chen.
\newblock Determination of optimal contract capacities and optimal sizes of
  battery energy storage systems for time-of-use rates industrial customers.
\newblock {\em IEEE Transactions on Energy Conversion}, 10(3):562--568, 1995.

\bibitem{Maly95}
D.~K. Maly and K.~S. Kwan.
\newblock Optimal battery energy storage system {(BESS)} charge scheduling with
  dynamic programming.
\newblock {\em IEE Proceedings - Science, Measurement and Technology},
  142(6):453 --458, 1995.

\bibitem{Monteiro13}
C.~Monteiro, T.~Santos, L.~A. Fernandez-Jimenez, I.~J. Ramirez-Rosado, and
  M.~S. Terreros-Olarte.
\newblock Short-term power forecasting model for photovoltaic plants based on
  historical similarity.
\newblock {\em Energies}, 6(5):2624--2643, 2013.

\bibitem{neufeld1987price}
J.~L. Neufeld.
\newblock Price discrimination and the adoption of the electricity demand
  charge.
\newblock {\em The Journal of Economic History}, 47(03):693--709, 1987.

\bibitem{nguyen2014optimal}
M.~Y. Nguyen and Y.~T. Yoon.
\newblock Optimal scheduling and operation of battery/wind generation system in
  response to real-time market prices.
\newblock {\em IEEJ Transactions on Electrical and Electronic Engineering},
  9(2):129--135, 2014.

\bibitem{nottrott2013energy}
A.~Nottrott, J.~Kleissl, and B.~Washom.
\newblock Energy dispatch schedule optimization and cost benefit analysis for
  grid-connected, photovoltaic-battery storage systems.
\newblock {\em Renewable Energy}, 55:230--240, 2013.

\bibitem{roberts2011role}
B.~P. Roberts and C.~Sandberg.
\newblock The role of energy storage in development of smart grids.
\newblock {\em Proceedings of the IEEE}, 99(6):1139--1144, 2011.

\bibitem{Romaus10}
C.~Romaus, K.~Gathmann, and J.~B{\"o}cker.
\newblock Optimal energy management for a hybrid energy storage system for
  electric vehicles based on stochastic dynamic programming.
\newblock In {\em Proceedings of the Vehicle Power and Propulsion Conference
  (VPPC)}, pages 1--6, Sept. 2010.

\bibitem{Rupanagunta95}
P.~Rupanagunta, M.~L. Baughman, and J.~W. Jones.
\newblock Scheduling of cool storage using non-linear programming techniques.
\newblock {\em IEEE Transactions on Power Systems}, 10(3):1279--1285, 1995.

\bibitem{sanghvi1989flexible}
A.~P. Sanghvi.
\newblock Flexible strategies for load/demand management using dynamic pricing.
\newblock {\em IEEE Transactions on Power Systems}, 4(1):83--93, 1989.

\bibitem{sheen1994time}
J.-N. Sheen, C.-S. Chen, and J.-K. Yang.
\newblock Time-of-use pricing for load management programs in {Taiwan Power
  Company}.
\newblock {\em IEEE Transactions on Power Systems}, 9(1):388--396, 1994.

\bibitem{smith2008advancement}
S.~C. Smith, P.~K. Sen, and B.~Kroposki.
\newblock Advancement of energy storage devices and applications in electrical
  power system.
\newblock In {\em Proceedings of the IEEE Power and Energy Society General
  Meeting}, pages 1--8. IEEE, 2008.

\bibitem{squartini2013optimization}
S.~Squartini, M.~Boaro, F.~De Angelis, D.~Fuselli, and F.~Piazza.
\newblock Optimization algorithms for home energy resource scheduling in
  presence of data uncertainty.
\newblock In {\em Proceedings of the International Conference on Intelligent
  Control and Information Processing (ICICIP)}, pages 323--328. IEEE, 2013.

\bibitem{taylor1986residential}
T.~N. Taylor and P.~M. Schwarz.
\newblock A residential demand charge: evidence from the {Duke} {Power}
  {Time-of-Day} pricing experiment.
\newblock {\em The Energy Journal}, 7(2):135--151, 1986.

\bibitem{Yoo12}
J.~Yoo, B.~Park, K.~An, E.~A. Al-Ammar, Y.~Khan, K.~Hur, and J.~H. Kim.
\newblock Look-ahead energy management of a grid-connected residential {PV}
  system with energy storage under time-based rate programs.
\newblock {\em Energies}, 5(4):1116--1134, 2012.

\bibitem{Yoon13-2}
Y.~Yoon and Y.-H. Kim.
\newblock An efficient genetic algorithm for maximum coverage deployment in
  wireless sensor networks.
\newblock {\em IEEE Transactions on Cybernetics}, 43(5):1473--1483, 2013.

\bibitem{Yoon13}
Y.~Yoon and Y.-H. Kim.
\newblock Geometricity of genetic operators for real-coded representation.
\newblock {\em Applied Mathematics and Computation}, 219(23):10915--10927,
  2013.

\bibitem{Yoon12}
Y.~Yoon, Y.-H. Kim, A.~Moraglio, and B.-R. Moon.
\newblock A theoretical and empirical study on unbiased boundary-extended
  crossover for real-valued representation.
\newblock {\em Information Sciences}, 183(1):48--65, 2012.

\bibitem{Youn09}
L.~T. Youn and S.~Cho.
\newblock Optimal operation of energy storage using linear programming
  technique.
\newblock In {\em Proceedings of the World Congress on Engineering and Computer
  Science Vol I}, pages 480--485, 2009.

\end{thebibliography}

\end{document}